\documentclass[letterpaper, 10 pt, conference]{ieeeconf}  % Comment this line out if you need a4paper

\IEEEoverridecommandlockouts                              % This command is only needed if 
\usepackage{graphics} % for pdf, bitmapped graphics files
\usepackage{graphicx,subfigure}
\usepackage{epsfig} % for postscript graphics files
\usepackage{amsmath} % assumes amsmath package installed
\usepackage{amssymb}  % assumes amsmath package installed
\usepackage{multirow} %multi row for the table
\usepackage{stfloats}

\title{\LARGE \bf PCA-aided Fully Convolutional Networks for \\
Semantic Segmentation of Multi-channel fMRI
}

\author{Lei Tai$^{1,3}$, Haoyang Ye$^{1,3}$, Qiong Ye$^2$ and Ming Liu$^{3}$% <-this % stops a space
\thanks{$^{*}$This paper is supported by Shenzhen Science, Technology and Innovation Comission (SZSTI) JCYJ20160428154842603 and JCYJ20160401100022706; partially supported by the HKUST Project IGN16EG12; partially supported by the Research Grant Council of Hong Kong SAR Government, China, under project No. 16206014 and No. 16212815, awarded to Prof. Ming Liu.}
\thanks{$^{1}$Lei Tai and Haoyang Ye are with MBE, City University of Hong Kong.
        {\tt\small \{lei.tai, hy.ye\}@my.cityu.edu.hk}}%
\thanks{$^{2}$Qiong Ye is with Department of Radiology, The First Affiliated Hospital of Wenzhou Medical University, Zhejiang, China.
        {\tt\small 94301699@qq.com}}%        
\thanks{$^{3}$Lei Tai, Haoyang Ye and Ming Liu are with Department of ECE, the Hong Kong University of Science and Technology. {\tt\small eelium@ust.hk}}%
}

\begin{document}

\maketitle
\thispagestyle{empty}
\pagestyle{empty}

%%%%%%%%%%%%%%%%%%%%%%%%%%%%%%%%%%%%%%%%%%%%%%%%%%%%%%%%%%%%%%%%%%%%%%%%%%%%%%%%
\begin{abstract}

Semantic segmentation of functional magnetic resonance imaging (fMRI) makes great sense for pathology diagnosis and decision system of medical robots. The multi-channel fMRI provides more information of the pathological features. But the increased amount of data causes complexity in feature detections. This paper proposes a principal component analysis (PCA)-aided fully convolutional network to particularly deal with multi-channel fMRI. We transfer the learned weights of contemporary classification networks to the segmentation task by fine-tuning. The results of the convolutional network are compared with various methods e.g. k-NN. A new labeling strategy is proposed to solve the semantic segmentation problem with unclear boundaries.
Even with a small-sized training dataset, the test results demonstrate that our model outperforms other pathological feature detection methods. Besides, its forward inference only takes 90 milliseconds for a single set of fMRI data. To our knowledge, this is the first time to realize pixel-wise labeling of multi-channel magnetic resonance image using FCN. %The related code and model will soon be released open-source.

\end{abstract}

%%%%%%%%%%%%%%%%%%%%%%%%%%%%%%%%%%%%%%%%%%%%%%%%%%%%%%%%%%%%%%%%%%%%%%%%%%%%%%%%
\section{Introduction}
%%%%%%%%%%%%%%%%Motivation%%%%%%%%%%%%%%%%%%%%
Medical-image analysis is indispensable in modern computer-aided diagnosis, therapy planning, and execution. Management of the related pathology relies on several different kinds of imaging processing technologies including positron emission tomography (PET), X-ray computer tomography (CT), ultrasound and magnetic resonance imaging (fMRI) \cite{barrah2016agent}. MRI is amongst the most useful technique for visualization. MRI shows high contrast and high resolution for different body tissues in a non-invasive way. Semantic segmentation of magnetic resonance image is an essential step for diagnosis with the abstraction of relevant features. %It addresses a number of significant problems. 
Through segmentation in the feature space, semantic segmentation minimizes the computational complexity in fMRI applications \cite{clarke1995mri}. For both diagnosis and treatment strategies, accurate classification and segmentation of the tissues are necessary and primordial, such as the quantitative volume measurement of different brain structures in brain fMRI segmentation \cite{huang2013automated}.

 %Two label strategy, and the change with training
   \begin{figure}[h!]
      \centering
      \includegraphics[scale=0.9]{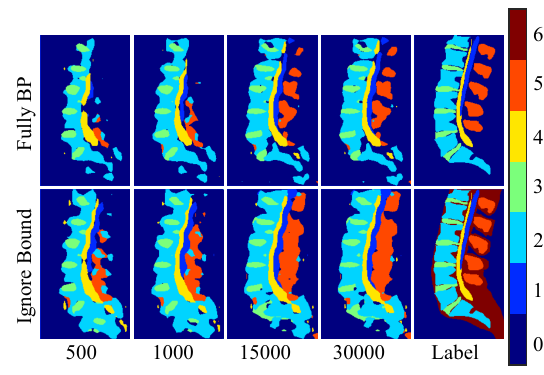}
      \caption{Representation of number: 0-back ground, 1-cerebrospinal, 2-vertebral bodyfluid, 3-lumbar disc, 4-spinal fluid, 5-bones, 6-ignore area. Different test result after 500, 1000, 15000, 30000 training iterations and the label instruction of two labeling strategies.}
      \label{fig:resultflow}
   \end{figure}

Convolutional networks have been applied to a variety of recognition and perception tasks recently, e.g. image classification \cite{simonyan2014very}, object detection \cite{ren2015faster} and semantic segmentation \cite{long2015fully}. The great success of Convolutional Neural Network (CNN) is attributed to the outstanding performance of trained hierarchical structure in data representations. The weights of CNN are typically calculated by back-propagation and gradient descent. This self-driven method has been keeping showing its unlimited potential in other conventional robotic fields \cite{lei2016deep} \cite{tl_rcar_2016} as well.

% Start from here
Pixel labeling, which is also called semantic segmentation, is much more challenging compared with image recognition and object detection. The scale of perception changes from a whole image to pixel-wise elements. For image classification, the coarse distribution is taken as the output. But for semantic segmentation, the output is a dense prediction at every pixel location.
Not like the traditional recognition and segmentation methods \cite{liu2012dp, liu2011regional, liu2012markov}, where mutual information is used manually, fully convolutional networks (FCN) \cite{long2015fully} is a CNN variant. It achieves the state-of-art effect in semantic segmentation task for most of the benchmarks. 

However, not like computer vision tasks with various datasets such as ImageNet \cite{russakovsky2015imagenet}, it is hard to collect an fMRI dataset for specific tissues of the human body and specific pathology features. Regular semantic segmentation dataset is comprised of 3-channel RGB images, while the parameters of fMRI are not constant based on the application and usage. Realizing such a segmentation in a small dataset with multi-channel fMRI means massive work load for both supervised expertise and medical diagnosis. The multi-channel infrastructure brings more co-related information, but also increases the data complexity. Training from small datasets like particular pathology and disease features can help build an expert system to recognize the same feature in the scene, which will bring great convenience for doctors. Another obstacle for pixel-wise segmentation in fMRI is the unclear boundary. Not like the typical RGB images by digital cameras, junction area between different tissues in fMRI are hardly recognizable even for experienced doctors. It also adds the complexity of the labeling process.

Motivated by the facts mentioned above, we stress the following contributions and features of this work:

\begin{itemize}

\item We propose a PCA-aided FCN structure to process 31-channel fMRI and to achieve pixel-wise semantic segmentation for primary tissues. Through PCA, the dimension of the origin data will be reduced to three and be further taken as the input of the fully convolutional network. 
\item The fine-tuning is implemented on a small-sized dataset as transfer learning. %Ming: check ``transfer learning'' to make sure it's true (Lei: Fine-tuning is a typical kind of transfer learning. But it seems that they both mean the same stuff here. http://cs231n.github.io/transfer-learning/).
The experiment results show outperforming efficiency compared with conventional methods for fMRI feature learning.
\item A revised labeling strategy is proposed, which ignores the junction areas between tissues, %Ming: by what?
which significantly decreases the complexity of semantic labeling for segmentation. 

\end{itemize}

%%%%%%%%%%%%%%%%%%Orginazation%%%%%%%%%%%%%%%%%%%
The rest of paper is organized as follows. We present related works in fMRI and pixel labeling in Section \ref{sec:rel}. In Section \ref{sec:pcafcn}, we describe our pre-processing configurations of the dataset with PCA and normalization. The configuration of the whole FCN structure is also introduced in this section. The detail of the training and evaluation on specific datasets are then presented in Section \ref{sec:experiment}. At the end, Section \ref{sec:conclusion} concludes the paper. 

%%%%%%%%%%%%%%%%%%%%%%%%%%%%%%%%%%%%%%%%%%%%%%%%%%%%%%%%%%%%%%%%%%%%%%%%%%%%%%%%
\section{Related Work} \label{sec:rel}

\subsection{fMRI Segmentation} \label{sec:mri_seg}
Most of conventional brain fMRI segmentation tasks are probabilistic, based on clustering methods like \textit{k-means}. Usually, the MR images are represented as a collection $X= \{\mathrm{x_1},\mathrm{x_2},\cdots, \mathrm{x_n} \}$ with every element representing a pixel of the image. The clustering algorithm returns $C= \{\mathrm{c_1},\mathrm{c_2},\cdots, \mathrm{c_m} \}$ as the cluster centers. $C$ are commonly initialized in random.

\textit{Barrah et al.} \cite{barrah2016agent} used a revised \textit{fuzzy clustering} algorithm to segment fMRI. Standard fuzzy c-means considers the clustering an optimization problem. The objective function is defined as:
%\operatorname{arg\,max}_a f(a) 
\[    \operatorname*{arg\,min}_C  \sum_{i}^{n} \sum_{j}^{m} w_{ij}^h {\| \mathrm{x_i} -\mathrm{c_j}\|}^{2}          \]
where $w_{ij}$ is the degree to which element $x_i$ belongs to cluster $c_j$ and $h$ is the fuzzifier to determine the level of cluster fuzziness. Euclidean distance is used to evaluate the element and cluster center directly. After that, $w_{ij}$ and $c_j$ can be estimated iteratively, as
\[      w_{ij} = \frac{1}{\sum_{k=1}^{m} \bigg( \big\|x_i-c_j  \big\| \big/   \big\|x_i-c_k  \big\|  \bigg)^{\frac{2}{m-1}} }     \]
\[  c_i = \sum_{j=1}^{m} w_{ij}^{h} x_j /  \sum_{j=1}^{m} w_{ij}^{h}    \]

In \cite{barrah2016agent}, the nearby area of the element in an image was considered as well. \textit{Gaussian mixture model} was also used to address brain MR images clustering \cite{huang2013automated}. Compared with \textit{fuzzy clusterng} mentioned above, the Gaussian kernel takes the place of norm functions between elements and cluster centers.

There are some challenges about ventricle fMRI segmentation, %Ming: what is ``Left ventricle....(% Lei: Sounds weird, but this is truly the name for the task)''
 such as Statistical Atlases and Computational Modeling of the Heart (STACOM) \cite{suinesiaputra2014collaborative} and  Medical Image Computing and Computer Assisted Intervention (MICCAI) \cite{radau2009evaluation}. Petitjean supplied a review of conventional methods for left ventricle segmentation\cite{petitjean2011review}, which consisted of graph cuts and deformable models, etc.

All of these methods provided a reasonable segmentation result for fMRI datasets. However, these statistical-based or feature-based method were limited, because they tend to overfit on specific dataset like ventricle. Their utility in a generic dataset was extremely limited as well.  

\subsection{ConvNet in Semantic Segmentation }
The convolutional neural network is leading the artificial intelligence right now. Related research \cite {silver2016mastering} just beat the best human player in Go game which was once regarded as the most difficult task for artificial intelligence. In image classification, deep residual network \cite{he2015deep} with hundreds of convolutional layers drove the network to learn identity feature and improve the accuracy of classification tasks in ImageNet \cite{russakovsky2015imagenet}. Faster-rcnn \cite{ren2015faster} addressed all of the tasks in object detection including region proposal, bounding box regression, instance location, and classification. CNN related methods take the state-of-the-art place in almost every computer vision tasks. Not only in vision fields, the successes of RNN and LSTM in natural language processing \cite{kalchbrenner2014convolutional} and speech recognition \cite{abdel2012applying} are all based on revised convolutional neural network. In terms of robot control, motivated by the game controller developed by Google \cite{mnih2015human}, deep reinforcement learning is implemented to teach the robot be adapted to the unfamiliar environment and achieve obstacle avoidance ability \cite{lei2016deep} \cite{tl_rcar_2016}. We adopt most of these concepts in this work.

Among various vision tasks, a high-level prediction like semantic segmentation is much more complicated compared with simply predicting image context as single output. While for pixel-wise classification, the computation source is thousands times more than the image classification.

Long \textit{et al.} \cite{long2015fully} re-architected and fine-tuned classification nets to achieve dense prediction of semantic segmentation directly. They improve the efficiency of convets in semantic segmentation dramatically because FCN extended a convolutional network to arbitrary-sized inputs \cite{matan1991multi}. Deep Convolutional Neural Networks (DCNN) were not sufficient to achieve semantic segmentation separately because of the poor localization property. Chen \cite{chen2014semantic} combined the feature map at the final layer of DCNNs with a fully connected Conditional Random Field (CRF). The main breakthrough in \cite{chen2014semantic} was the intersection of convolutional network and probabilistic graphical models. Notice that, CRF and FCN are the state-of-the-art semantic segmentation methods nowadays. \cite{tran2016fully} applied FCN processing cardiac segmentation in the ventricle dataset mentioned in Section \ref{sec:mri_seg}, which also focused on a unique dataset but not generic multi-channel fMRI. %Ming: what is the main difference to {tran2016fully}?

%%%%%%%%%%%%%%%%%%%%%%%%%%%%%%%%%%%%%%%%%%%%%%%%%%%%%%%%%%%%%%%%%%%%%%%%%%%%%%%%
\section{PCA-aided Fully Convolutional Networks} \label{sec:pcafcn}
Our approach benefits a lot from the recent popular deep net-related works including image classification \cite{simonyan2014very}, semantic segmentation \cite{long2015fully} and transfer learning \cite{bengio2012deep}. After PCA, the processed fMRI data are taken as the input of FCN.

\subsection{Data Gathering and Preparation} \label{sec:data_gather}
%data gathering procedure
This study was approved by the local institutional ethic review board. Written informed consent was obtained from all volunteers. Six young and healthy volunteers
(age range, 40 years; median age, 38 years; three males and three females)
were recruited in this study. In addition to standard criteria for exclusion from fMRI, further exclusion criteria were applied using the absence of cervical spine related pain and disorders. 

The dataset focuses on the vertebral area of human being. Except the background, there are five primary tissues including bones, spinal fluid, lumbar disc, vertebral body and cerebrospinal fluid shown in Fig. \ref{fig:resultflow}.

fMRI examinations were carried out on 3T clinical MR scanner (Achieva, Philips Healthcare, Best, The Netherlands) with standard abdomen coil. 
Mid-sagittal multi-echo T2-weighted images were acquired using a multi-shot spin-echo sequence with echo-planar readout and following parameters shown in Table \ref{tab:table_1}. Here \textit{TE} means echo time. The intensity of image under different echo time reflects the attenuation of tissues. With 32 different echo times, 32 channels representing the different intensity of various tissues. The first echo of the sequence was excluded from further analysis to avoid the effect of the stimulated echo \cite{takashima2012correlation}. After cutting the black background edge, the dimension of every sample in the dataset turns to be $31\times256\times154$.

\begin{table}
    \centering
    \caption{fMRI gathering parameters}
    \label{tab:table_1}
    \begin{tabular}{c c c}
    \hline
    \hline
    Parameter & Value & unit \\
    \hline
    FOV    &    $240\times155\times5 (FH \times AP \times RL)$ & $mm^3$ \\
    matrix & $120\times72$ & - \\
    reconstruction & $32$ & - \\
    slice thickness & $5.0$ & $mm$\\
    No. of slice & $1$ & -\\
    No. of TEs & $32$ & -\\
    TE/TR & $8.0\times32/1202$ & $ms$\\
    NA & $8$ & -\\
    fat suppression & SPIR & -\\
    scan time & $235.5$ & $s$\\
    \hline
    \end{tabular}
\end{table}

\subsection{PCA Processing} \label{sec:label_strategy}
It is evident that the 31-channel fMRI provide more information in detail about the pathology features. But because 31-channel images increase the calculation complexity dramatically, a dimension reduction method is necessary. Besides, state-of-the-art CNN structures were usually implemented on vision tasks which take raw RGB images with three channels as input. Preliminaries experiments also tell that if we attach the images with 31 channels as input layer to the network, the weights of the network is very difficult to converge. These difficulties are to be moderated.
%%%%%%%Why PCA Make sense%%%%%%
%%Figure of orgin / pca
   \begin{figure}[h]
      \centering
      \includegraphics[scale=1]{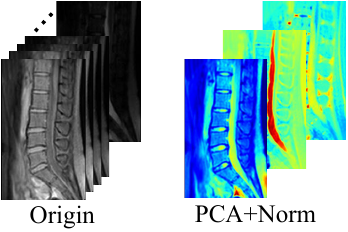}
      \caption{Data transformation from 31 channels to 3 channels after PCA processing. Origin data is collected from 6 volunteers. The dataset shows vertebral area of human beings.}
      \label{fig:pcafigure}
   \end{figure}

In PCA analysis, we vectorize every channel of the sample to a vector. The sample is transformed to a matrix $X$ as $31\times39424$. PCA is a variant matrix transformation as

\[
    \hat{X}_k =X - \sum_{s=1}^{k-1} X w_{(s)} w_{(s)}^{T}
\]

Here $w_(s)$ means the related eigenvector of $XX^T$'s sorted eigenvalues. And $w_(s)$ is also a united vector. It can be calculated like
\[
    w_{(1)} = \arg \max \bigg(  \frac{w^{T}X^T X w}{w^T w}   \bigg) 
\]

\begin{figure}[h]
      \centering
      \includegraphics[scale=0.6]{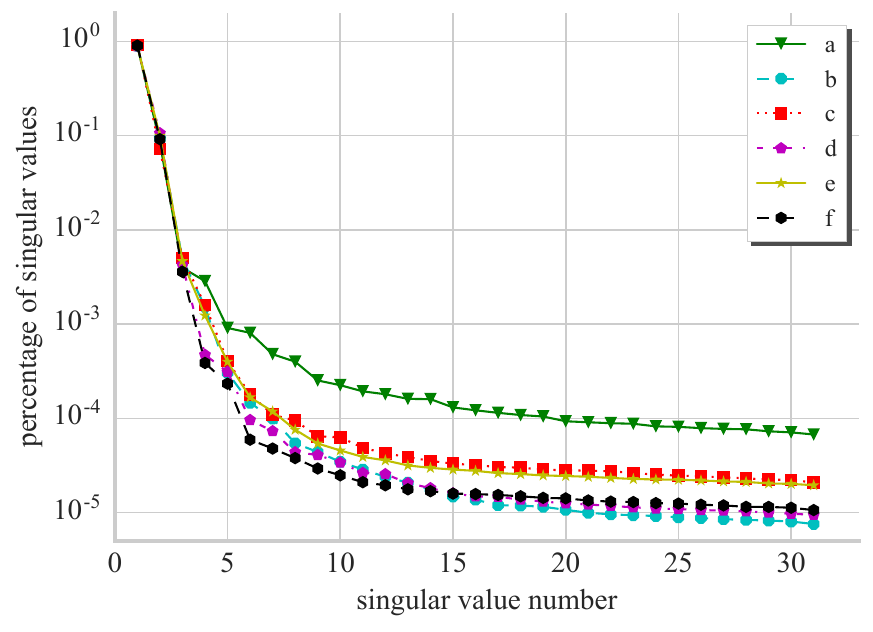}
      \caption{Percentage of singular values of all the samples in datasets. Singular values are sorted and the percentage is shown in $log$ scale.}
      \label{fig:pcaresult}
   \end{figure}

The left part of Fig. \ref{fig:pcafigure} shows a sample of the origin 31-channel images and the transformation result after PCA and normalization processing. Fig. \ref{fig:pcaresult} shows all of the 31 singular values of $X$ which is also the square root of eigenvalues of $XX^T$ from all of the six samples of the datasets. The first three singular values take the weights of all the singular values in more than $99.9\%$. That means the transformation is more efficient and meaningful. The largest three singular values of $X$ can almost represent all of the origin information in the raw sample equivilently.
   
The fMRI datasets have a broad range of pixel intensities. The different ranges significantly influence the accuracy of segmentation. Even after PCA processing, the ranges of the three channels are still quite irregular. To locate the maximal and minimal elements among the three channels, we normalize the three channels to a range of (0,255), which is the range of 8-bit images. The right part of Fig. \ref{fig:pcafigure} illustrate the sample condition after PCA and normalization. 
   \begin{figure*}[t!]
      \centering
      \includegraphics[scale=0.6]{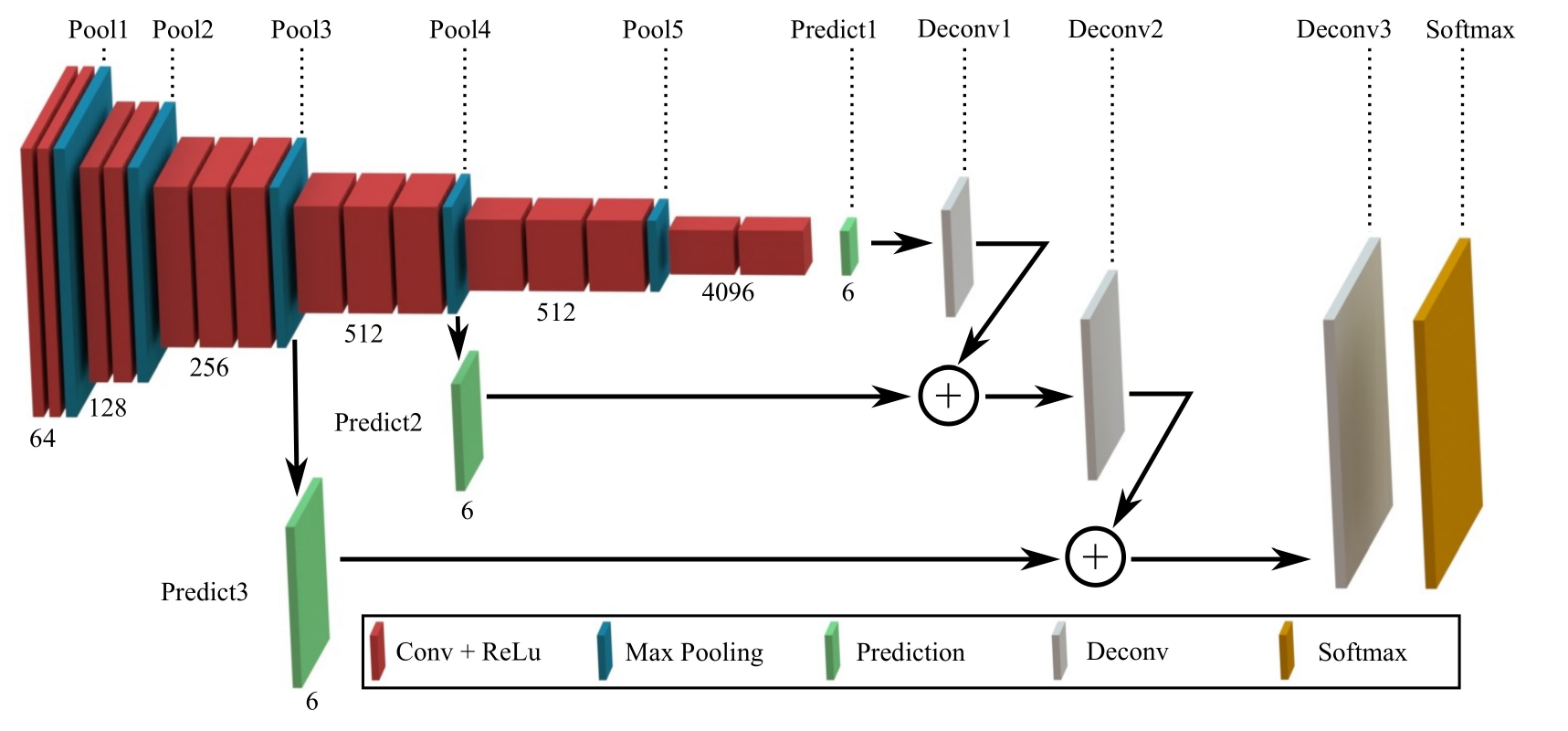}
      \caption{The structure of our proposed fully convolutional neural network. Feature learning part use the structure of \textit{VGG-16} net\cite{simonyan2014very}. After feature learning, the pooling output of different stages are tackled with upsampling and merged. The final layer uses a Softmax layer to output dense classification result matrix.}
      \label{fig:net_work}
   \end{figure*}
The supervising learning procedure will be introduced later in Section \ref{sec:sl_learing}. %Ming: which section? 
The labeling process is tackled on the origin sample before PCA pre-processing. We propose two different labeling strategies as shown in Fig. \ref{fig:resultflow}, dealing with five categories of primary tissues in the vertebral fMRI datasets, as follows:
\begin{itemize} %Ming: re-phrase these two items; not clear.
\item \textit{Fully-BP:} The origin image is aggressively segmented to six parts including all of the five primary classes as introduced in Section \ref{sec:data_gather} and the background area. The predictions in every pixel location are to be added in the loss function and used to execute back propagation for weights.
\item \textit{Ignore-Bound:} Among the primary tissues, the boundary (separation plane) is vague even for human inspectors. We ignore these boundary parts and label no targets for pixels located here. The forward predictions of these parts will not be considered when calculating the loss. Naturally, there is no back propagation through these locations.

\end{itemize}

\subsection{FCN Implementation} \label{sec:sl_learing}

Fig. \ref{fig:net_work} illustrates our proposed FCN architecture.  In the convolutional networks, each layer is a three-dimensional array of size $h \times w \times d$, where $d$ is the channel dimension and $h,w$ are the spatial sizes. Here the dimension of the input data is following the pre-processed fMRI data. The size of input array is thus $256 \times 154 \times 31$.

\subsubsection{Dense Prediction}
%%%%%%FCN multi structure%%%%%%
%Trainning dataset loss decreased
The feature learning of FCN uses the structure of \textit{VGG-16} net \cite{simonyan2014very} which won ILSVRC14. As introduced in \cite{long2015fully}, all of the fully connected layers of \textit{VGG-16} are converted to convolutions. Finally, there are 15 convolutional (Conv) layers for feature learning. To increase non-linearity, every Conv layer is followed by a Rectified Linear Unit (ReLU) activation function layer. They are represented as red cubes in Fig. \ref{fig:net_work}. The number under the Conv+ReLU cubs in Fig. \ref{fig:net_work} is the number of channels of the output data related to this cube. A $1 \times 1$ convolution layer with channel dimension six is appended to predict scores for each of the tissue classes including the background. As shown in Fig. \ref{fig:net_work}, actually the Prediction cube is also a convolution layer. All of the output channel numbers of prediction convolution layers is six as well. The Deconv layer will upsample the coarse output to the similar size as input ($256 \times 154$).  
\begin{table}
    \centering
    \caption{Training parameters}
    \label{tab:table_3}
    \begin{tabular}{c c}
    \hline
    \hline
    Parameter & Value \\
    \hline
    batch size    &    $1$\\
    learning rate &  $10^{-14}$ \\
    momentum    &    $0.99$ \\
    weight decay  &  $0.0005$\\
    \hline
    \end{tabular}
\end{table}
\subsubsection{Hierarchy Combination}
Max pooling layers, represented by blue cubes, are implemented after several Conv+ReLU cubes to spatially reduce the data size. We use the same strategy to combine layers of the feature hierarchy as introduced in \cite{long2015fully}. The output of \textit{Pool3}, \textit{Pool4}, \textit{Pool5} are to be unsampled to with different strides. As the sequence illustrated in Fig. \ref{fig:net_work}, \textit{Deconv1} and \textit{Predict2} will be firstly added conjugatedly and unsampled to \textit{Deconv2}. Then \textit{Deconv2} and \textit{Predict3} will be added in sequence and unsampled to \textit{Deconv3}. The dimension of \textit{Deconv3} is $6 \times 256 \times 154$. At last, a Softmax layer is appended to output the final prediction.

%%%%%%%%%%%%%%%%%%%%%%%%%%%%%%%%%%%%%%%%%%%%%%%%%%%%%%%%%%%%%%%%%%%%%%%%%%%%%%%%
\section{Experiments and Results} \label{sec:experiment}

\subsection{Training and Test}
The whole structure is trained end-to-end and the initial weights of network are transferred from the training result of VOC dataset \cite{everingham2010pascal} in \textit{voc-fcn8s} \footnote{http://dl.caffe.berkeleyvision.org/fcn8s-heavy-pascal.caffemodel} structure which was pre-trained with the parameters in \cite{long2015fully}. In the fine-tuning procedure, we use Stochastic Gradient Descent (SGD) method with momentum. The latest research \cite{DBLP:journals/corr/ShelhamerLD16} showed that setting the batch size to one is beneficial for semantic segmentation. We tried different learning rate from $10^{-5}$ to $10^{-14}$, and smallest learning rate results in the best performance. One possible explanation is that the initialization from ImageNet-trained weights is important, so the tuning that makes it adaptive to the new datasets should be tiny and slow. Training parameters is described in detail in \ref{tab:table_3}.

We use Caffe \cite{jia2014caffe}  \textit{python} interface as the deep learning framework to train and test all of the data and models. A unit NVIDIA Tesla K40c is used to support the computation.

 %Trainning dataset loss decreas
   \begin{figure}[h]
      \centering
      \includegraphics[width=0.95\columnwidth]{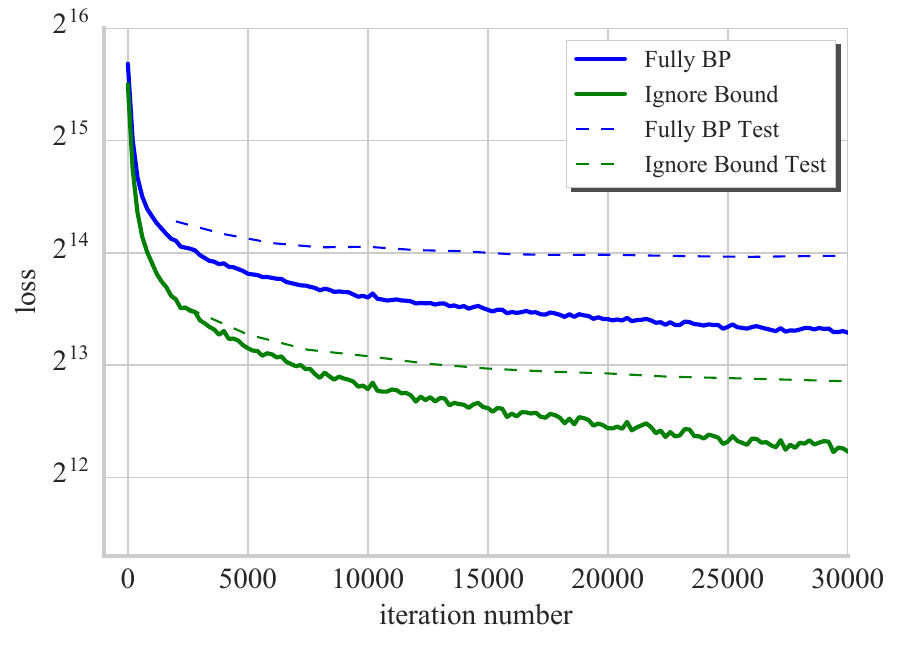}
      \caption{The decrease process of training and test loss of two different label strategies. The multi-nominal logistic loss of all of the pixels in the sample image are added.}
      \label{fig:loss_decrease}
   \end{figure}

There are six MR images in the vertebral datasets as illustrated in Fig. \ref{fig:pcaresult}. We separate one of them as the test sample. Fig. \ref{fig:loss_decrease} shows the training and test loss decreasing with the iteration number increasing. 
The two strategies introduced in Section \ref{sec:label_strategy} are all implemented and tested.
All of the losses experience substantial reduction at first 5000 steps. After that, because of the existence of ignored area, the loss of \textit{Ignore-Bound} strategy is obviously less than the one with the \textit{Fully-BP} strategy. The test losses of both strategies are converged stably after 15000 iterations. 

\subsection{Results}

\begin{figure*}[!ht]
    \centering
    \subfigure[Fully-BP]{\includegraphics[width=0.45\textwidth]{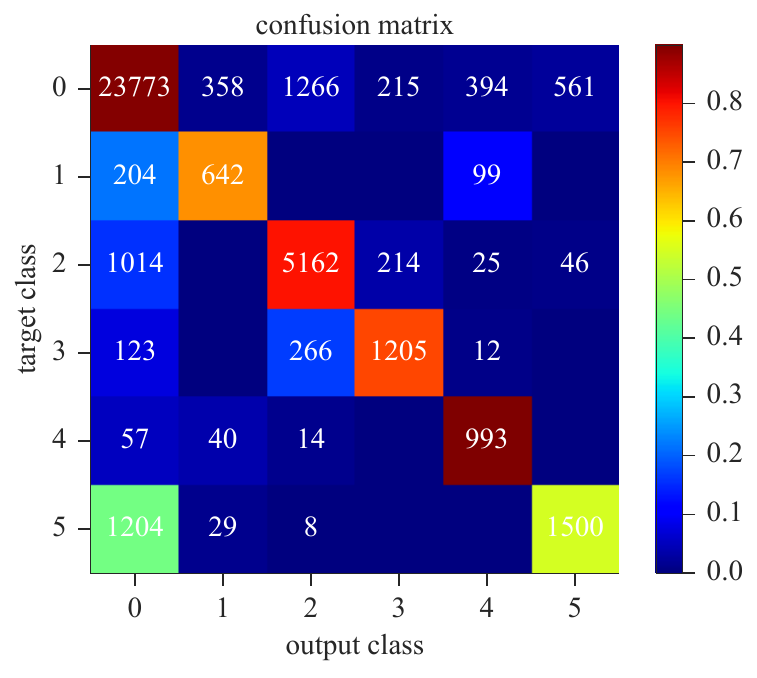}
    \label{fig:cfmatrix_full}}
    \subfigure[Ignore-Bound]{\includegraphics[width=0.45\textwidth]{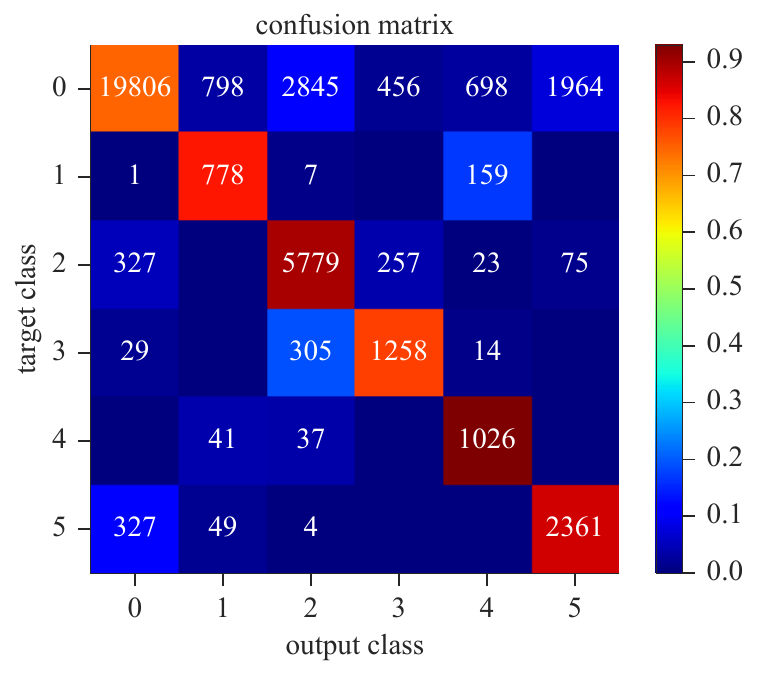}
    \label{fig:cfmatrix_class_7}}
    \caption{The confusion matrix of test results with two strategies after 15000 iterations of training. Representation of number: 0-back ground, 1-cerebrospinal, 2-vertebral bodyfluid, 3-lumbar disc, 4-spinal fluid, 5-bones. The number in the grid located at row $i$ and column $j$ means pixels in the test sample labeled with class $i$ and predicted to belong to class $j$, which is $n_{ij}$ introduced in Table \ref{tab:table_critteria}. The color of this grid represents the precision ($n_{ij}/s_i$). The number of $n_ii$ on the diagonal is also called true positive count.}
%Ming: re-phrase the caption. Some sentences are unclear, say: which is the ratio of the number to the sum of this row.
    \label{fig:cf_matrix_all}
\end{figure*}

%Test dataset results confusion matrix 
    
%Compare with the other method, two different training strategy

A naive K-Nearest-Neighbor (KNN) method, which is commonly used nowadays, is implemented as comparison as well. 
%We calculate the medians of all the different classes including the background of all the training samples after PCA and normalization process. 

%Ming: re-phrase the following:
In this experiment, there are five samples in the training dataset as mentioned in Section \ref{sec:data_gather}. From the training dataset, we got $5 \times 256 \times 154$ pixels for each scan. After PCA and normalization process, there is a 3-dimensions vector for every pixel location. We calculate the median 3-dimensional vectors for pixels belong to same class and denote it by $y_j$ for class $j$. For all of the six classes introduced in Section \ref{sec:data_gather}, let $Y=\{y_1,y_2,\dots y_6 \}$ be the medians set for all of six classes. For a pixel in the test sample after PCA and normalization process, let the 3-dimensional pixel vector be $x$. We can locate the least normalized inner product between $x$ and the element in the median sets $Y$ as:
\[    
\operatorname*{arg\,min}_j  \frac{x^T  y_j}{\|x\| \|y_j\|}
\]
Then $j$ is the prediction result of pixel $x$. By scanning the whole image, we can get the prediction for every pixel in the test data.
%Ming: ^ to here. 

\begin{table}[!h]
    \centering
    \caption{ Parameter Representations for criteria}
    \label{tab:table_critteria}
    \begin{tabular}{c | c c }
    \hline
    \hline
    Parameters & Definitions\\
    \hline
     $i,j$ &  specific class number\\
     $n_{ij}$ & pixels of class $i$ predicted to belong to class $j$\\
     $n_c$  & number of classes (6)  \\
     $s_i$ & pixels labeled with class $i$ ($\sum_j n_{ij} $)\\
    \hline
    \end{tabular}
\end{table}

 \begin{table*}[!ht]
    \centering
    \caption{Test results for FCN based on two strategies at the related training section and the naive method. }
    \label{tab:table_2}
    \begin{tabular}{l | c c c c| c c c c  }
    \hline
    \hline
   \multirow{2}{*}{Methods}  &  \multicolumn{4}{|c|}{all classes}  & \multicolumn{4}{|c}{main tissues} \\
   & mean IU & fw IU& pixel acc & mean acc & mean IU & fw IU& pixel acc & mean acc \\ \cline{1-1}
    \hline
    Naive KNN         & 13.1  &  13.0  & 23.0 &  44.6 & 13.5  & 16.4  &  44.5 & 38.3 \\
    Fully-BP 10000  & 58.8  &  73.3 &  83.9 & 75.8  & 54.4  & 57.3  & 74.1  & 80.1 \\
    Fully-BP 15000  & 59.6  &  73.8 &  84.4 & 76.2 &  55.2 &  58.0  & 73.9 & 80.5\\
    Fully-BP 30000  & \textbf{60.7} & \textbf{74.4} & \textbf{84.7} & 78.3 & \textbf{56.5} & \textbf{59.3} & 75.1 & 81.7 \\
    Ignore-Bound 10000 & 54.8 & 66.5 & 78.4 & \textbf{84.2} & 54.4 & 54.5 & 87.1 & \textbf{83.8} \\
    Ignore-Bound 15000 & 55.0 & 66.9 & 78.7 & 84.0 & 51.5 & 54.9 & 87.1 & 83.5 \\
    Ignore-Bound 30000 & 54.1 & 66.3 & 78.2 & 83.8 & 50.6 & 54.6 & \textbf{87.6} & 82.8  \\
    \hline
    \end{tabular}
\end{table*}

We use several common semantic segmentation and scene parsing methods to compare the experiment results, using criteria like pixel accuracy and region intersection over union (IU). Based on the parameters introduced in Table \ref{tab:table_critteria}, we compute:

\begin{itemize}
\item mean IU: \\
\[(1/n_c) \sum_i n_{ii} /(s_i + \sum_j n_{ji} - n_{ii} ) \]
\item frequncy weighted IU (fw IU): \\
 \[ {(\sum_k s_k)}^{-1} \sum_i s_i n_{ii}/(s_i + \sum_j n_{ji} - n_{ii} ) \]
\item pixel accuracy: \\
\[ \sum_i n_{ii}/\sum_i s_i \]
\item mean accuracy: \\
\[ (1/n_c)\sum_i n_{ii}/s_i \]
\end{itemize}

For the particular target of segmentation for medical images, the background class is not as essential as the other meaningful tissues. When we calculate the loss in the back propagation procedure of training, the same metrics for all of the five primary tissues except the background are also introduced as shown in \ref{tab:table_3}.

The evolution of test results and label maps are described in Fig. \ref{fig:resultflow}. As the result of loss reduction, the shapes of the segmentation results progressively tended to the same as the ground-truth at the first several thousands steps. After 15000 iterations, %Ming: ``steps'' or ``epochs'' Lei: it's the same. Cause the batch number is 1
there is hardly any revision anymore. By comparison, we find that the \textit{Fully-BP} strategy prefers to predict the junction area as background. On the other hand, the \textit{Ignore-Bound} strategy is prone to regard the junction area as tissues especially for bone area. The confusion matrices of the test result after 15000 iterations for both strategies which reflect the difficulties of prediction for bones are illustrated in Fig. \ref{fig:cf_matrix_all}. An probable explanation is that the bone area is the most difficult one to label with unclear boundary mostly as shown in Fig. \ref{fig:pcafigure}. There may be too many labeling mistakes in this area with the \textit{Fully-BP} strategy. For the precision of different classes, the \textit{Ignore-Bound} strategy substantially improves the precision of all classes except the background part. However, it takes the cost to increase the false positive counts for primary tissues as well.
%Ming: why bone are is difficult.  Tai: unclear boundary motstly. Just add it.
%Ming: how about a ROC or Recall-precision curve? Tai: the last two sentences is trying to explain that the recall is a littlt worse for the second one. R-P curve is based on the classification solver with a threshould, is it appropriate for muclti class CNN?

Comparison among various methods and FCN methods against different training periods are listed in Table \ref{tab:table_3}. FCN methods show great improvement compared with other methods. Another fact is that FCN only needs 90 milliseconds to calculate the forward prediction, and the conventional method takes 1.4 seconds to finish the traversal analyzing for every pixel in the whole image. %Ming: while (other methods) require how much time? %1.4s is for the navie method, the explanation is not clear.
From Table \ref{tab:table_3}, the \textit{Fully-BP} strategy shows high performance in mean-IU and frequency-weighted-IU. %Ming: <- not a sentence. 
The \textit{Ignore-Bound} strategy has increased accuracy especially for primary tissues. For the five primary classes of tissues except the background, the differences in mean IU and fw IU are also reduced compared with the \textit{Fully-BP} strategy. %Ming: what is IU? %Ming: <- re-phrase; not clear and duplicate the previous. %TAI: Definition of IU has been introduced in the metrics part. 
And the best performance of the \textit{Ignore-Bound} strategy appeared after around 10000 iterations. It takes less time to achieve the satisfied result compared with the \textit{Fully-BP} strategy which need 30000 iterations. Generally speaking, the \textit{Ignore-BP} strategy shows benefits in converging speed and accuracy.  %Ming: This paragraph is very badly written with unclear justification. It's ok to use multiple lengthy paragraphs, as long as it's explicit. Please re-write this paragraph. 
 
%%%%%%%%%%%%%%%%%%%%%%%%%%%%%%%%%%%%%%%%%%%%%%%%%%%%%%%%%%%%%%%%%%%%%%%%%%%%%%%%%%
\section{Conclusion} \label{sec:conclusion}
%can be improved
This papers demonstrated the effectiveness to apply FCN deep-learning methods on semantic segmentation of multi-channel fMRI. The utility of a PCA-aided fully convolutional network model is proved under efficient end-to-end training using a small-sized dataset. It indicates that the transfer learning and \textit{Ignore-Bound} label strategy are evidentially sufficient when collecting small datasets, labeling them and realizing the pixel-wise classification in near real-time. The proposed framework outperforms other state-of-the-art methods in precision. 

A few limitations to our approach remain. Most notably, as with most CNN approaches, the training parameters is decided after a lot of attempts. Fine-tuning from other pre-trained models is also time consuming. The labeling process of \textit{Ignore-Bound} strategy need an appropriate threshold to distinguish the junction area.

In the future, the labeling strategy can be further improved considering the limitations in parts of the experimental results. The combination with CRF introduced in \cite{chen2014semantic} should also be implemented and tested. Eventually, we will not only test on the tissue segmentation but also apply the segmentation method on pathology related areas directly. %Ming: also list some shortcomings of the proposed method to protect yourself from reviewers' attack.

%%%%%%%%%%%%%%%%%%%%%%%%%%%%%%%%%%%%%%%%%%%%%%%%%%%%%%%%%%%%%%%%%%%%%%%%%%%%%%%%
%\section*{Acknowledgment}


%%%%%%%%%%%%%%%%%%%%%%%%%%%%%%%%%%%%%%%%%%%%%%%%%%%%%%%%%%%%%%%%%%%%%%%%%%%%%%%%
\bibliographystyle{IEEEtran}
\bibliography{MRI_bib}

\end{document}